\pdfoutput=1 

\documentclass{article} 
\usepackage{iccv}
\usepackage{times}
\usepackage{epsfig}
\usepackage{graphicx}
\usepackage{amsmath}
\usepackage{amssymb}
\usepackage{subfigure}
\usepackage{tabulary}
\usepackage[section]{placeins} 
\usepackage{array}
\usepackage{booktabs}
\newcommand{\tld}{\raise.17ex\hbox{$\scriptstyle\mathtt{\sim}$}} 

\iccvfinalcopy 

\ificcvfinal\fi
\begin{document}


\title{Shallow Networks for High-Accuracy Road Object-Detection}

\author{Khalid Ashraf, Bichen Wu, Forrest N. Iandola, , Mattthew W. Moskewicz, Kurt Keutzer\\
Electrical Engineering and Computer Sciences Department, UC Berkeley\\
{\tt\small \{ashrafkhalid, bichen\}@berkeley.edu, \{forresti, moskewcz, keutzer\}@eecs.berkeley.edu }
}

\maketitle
\vspace{-0.1in}
\begin{abstract}
The ability to automatically detect other vehicles on the road is vital to the safety of partially-autonomous and fully-autonomous vehicles. Most of the high-accuracy techniques for this task are based on R-CNN or one of its faster variants. In the research community, much emphasis has been applied to using 3D vision or complex R-CNN variants to achieve higher accuracy. However, are there more straightforward modifications that could deliver higher accuracy? Yes. We show that increasing input image resolution (i.e. upsampling) offers up to 12 percentage-points higher accuracy compared to an off-the-shelf baseline. We also find situations where earlier/shallower layers of CNN provide higher accuracy than later/deeper layers. We further show that shallow models and upsampled images yield competitive accuracy. Our findings contrast with the current trend towards deeper and larger models to achieve high accuracy in domain specific detection tasks. 

\end{abstract}
\vspace{-0.1in}

\section{Introduction and Motivation}
\label{sec:intro}

\label{sec:intro}

Advanced driver assistance systems (ADAS) and increasingly autonomous vehicles promise to make transportation safe, efficient and cost effective. Driven by the goal of building a safe transportation system, ADAS has emerged as a leading research direction in recent years ~\cite{hillel2012,huval2015,rajpurkar2015,deepDriving}. Some specific ADAS related machine learning tasks include detection of road boundaries, lane topologies, location of other cars, pedestrians, road signs and obstacles. These detection capabilities form the core of an ADAS technology stack. Other parts of the technology stack include decision making and control systems that take action in a certain road situation based on the input from the perception system. Although this picture works in controlled environments, making this technology effective in changing road situations, emergencies, changing weather etc. remains a significant challenge. 

In recent times, deep learning has shown leading accuracy in a number of machine learning challenges. Specifically relevant to ADAS application is the dramatic increase in accuracy of image object classification~\cite{alexnet,googlenet,VGG-19,resnet} and localization~\cite{overfeat,segDeepM,R-CNN,Fast-R-CNN,Faster-R-CNN} in the last few years. A key advantage of DNN-based approaches is that they do not require hand tuned features for detecting every object but rather learn the representation from the data itself. Deep learning based perception systems promise to play a key role in navigation and safety software stack for ADAS. 

R-CNN and its faster variants have become the state of the art in different object detection tasks. In this work, we leverage this method to establish a number of observations related to car detection on the challenging KITTI ~\cite{kittiData} dataset. Our main results can be summarized as:

\begin{itemize}
\item \textbf{Bigger input images lead to higher accuracy.}
Input image resolution increases the accuracy of car detection using the faster R-CNN network.

\item \textbf{Shallow models can deliver high accuracy.}
Convolutional features from shallow or earlier layers of DNNs lead to higher accuracy than features from the deeper layers. This holds true for deep models like VGG16. Surprisingly, even shallow models like AlexNet provide high accuracy on the detection task. 
Using shallow models that require less memory allow us to use very high input image resolutions.
In terms of accuracy, shallow models with high resolution are competitive with deeper models with traditional resolution.
This result is surprising given the trend of searching for deeper models for achieving high accuracy on object detection tasks.

\end{itemize}

The rest of the paper is organized as follows.
In Section~\ref{sec:related} we review related work, and we provide technical background information in Section~\ref{sec:preliminaries}.
We describe our initial experimental setup in Section~\ref{sec:experimental-setup}.
We get to the crux of our results about large image resolution in Section~\ref{sec:image_resolution} and shallow models in Section~\ref{sec:shallow_models}.
We do additional exploration of R-CNN based configurations in Section~\ref{sec:further-improvements}
We summarize our findings in the context of the related work in Section~\ref{sec:discussion}, and we conclude in Section~\ref{sec:conclusions}.

\section{Related Work}
\label{sec:related}

\subsection{Deep Networks for Object Detection}

Deformable parts models (DPM) were the state of the art for image object detection~\cite{FelzenszwalbPAMI10} before the emergence of deep convolutional neural nets. 
The R-CNN method uses selective search for object region proposal~\cite{R-CNN}. The proposed regions in an image are warped to a fixed size and fed into a classification network called R-CNN. Fast R-CNN was introduced to reuse the shared convolution features for the region proposals ~\cite{Fast-R-CNN}. In {\em Fast R-CNN}~\cite{Fast-R-CNN}, the inference speed is still dominated by the region proposal in the selective search method. {\em Faster R-CNN}~\cite{Faster-R-CNN} proposes object bounding boxes directly from the convolutional features. Inspired by the SPPNet~\cite{SPPnet} method, Faster R-CNN uses a region proposal network (RPN) to regress proposal boxes to ground truth boxes. The regions proposed by the RPN network is fed into the R-CNN network for classification. The network is trained end to end.\footnote{The Faster R-CNN codebase also offers piecewise training of RPN and classifier branches of the network, but we found this cumbersome, and we use end-to-end training in all of our Faster R-CNN experiments.}  
Other than the RPN based method, there are several methods proposed for object bounding box prediction.
For example, the OverFeat~\cite{overfeat} method predicts a single box for localization whereas the Multibox~\cite{multibox-1,multibox-2} method predicts multiple boxes in a class-agnostic way. The SPP  method ~\cite{SPPnet} uses shared convolutional feature maps for fast object detection.

\subsection{Detection on the KITTI dataset}

\subsubsection{Detection using 2D data}

Deep neural networks are the backbone of most high-accuracy approaches to identifying objects such as cars in KITTI and similar datasets. 
Many such methods have been proposed; we focus this section on the high-accuracy and peer-reviewed results.
A high-accuracy method for identifying objects in KITTI dataset is scale dependent pooling (SDP) combined with cascaded region classifiers (CRC)~\cite{SDP_CRC}. 
The crux of SDP+CRC lies in selecting a high-resolution CNN layer (e.g. conv3\_3 in VGG16~\cite{VGG-19}) or a heavily downsampled CNN layer (e.g. conv5\_3), depending on the resolution of each region proposal.
By combining features from multiple convolution layers, they were able to achieve very high accuracy on KITTI's object detection task. 
Our method introduced in this paper is even simpler in that we use only a single layer for feature extraction. 

Another approach is {\em Monocular 3D (Mono3D)} which actually uses 2D images, but it aims to identify the pose of objects, with the goal of detecting objects as 3D bounding boxes.
Like SDP+CRC, Mono3D is built around a version of R-CNN.
There are also a number of anonymous and/or sparsely-explained submissions to the KITTI website's leaderboard that are reportedly built on top of R-CNN.

\subsubsection{Detection using 3D data}

The KITTI dataset provides 3D information in the form of stereo images and LIDAR point clouds.
Recent results such as 3DVP~\cite{xiang2015} and 3DOP~\cite{3DOP} leverage both 2D and 3D data to achieve higher accuracy relative to comparable 2D baselines.

To build supervised 2D datasets such as ImageNet~\cite{imagenet} and PASCAL~\cite{PASCAL}, a widely-used approach is to have mechanical turk workers annotate user-generated images and videos from websites such as Flickr or YouTube.
However, to our knowledge, there is no 3D equivalent of Flickr or YouTube that receives petabytes per week of user-uploaded 3D imagery.
As a result, the overhead in building a 3D dataset currently requires not only data annotation, but also {\em data collection}.
The cost of data collection includes hours of human labor, and it can also require expensive sensors.
The KITTI dataset was released several years ago. However, the Velodyne HDL-64E LIDAR scanner used by the KITTI team still costs 80,000 USD\footnote{http://articles.sae.org/13899}, which is more than twice the price of the average new car in the United States. 
With all of this in mind, we think widespread research on 3D object detection will be slow to emerge until (1) there is an internet hub that attracts large quantities of user-generated 3D imagery, and (2) the equivalent of today's high-end LIDAR sensors become available for tens of US dollars.
With this in mind, we focus our efforts on 2D imagery, where anyone with modest resources can collect a custom training set and apply our object detection approach.

\section{Preliminaries}
\label{sec:preliminaries}

\subsection{Relationship of conv and pooling strides to activation grid dimensions}

The feature dimension of the output of a spatial convolution operation depends on the dimension of its input and the strides used (ignoring the boundary effects). In convolutional neural nets, strides are used to continually reduce the feature dimensions in the convolution layers. Additionally pooling layers are used to reduce dimension by averaging or taking maximum value within a neighborhood. Thus the spatial dimension of the output of the convolution decreases as we go to deeper layers. For example, in VGG16, for a standard input image size 224x224, the features calculated by the first convolution layer is 224x224 which reduce to 14x14 at the output of conv5\_3. This dimension is further reduced by the pool5 layer to 7x7. Similarly for AlexNet, the convolution feature dimension reduces from 55x55 in conv1 to 6x6 in pool5.

In some CNN architectures such as AlexNet~\cite{alexnet} and VGG16~\cite{VGG-19}, the first fully-connected layer expects a specific height and width for its input data (e.g. 6x6 for AlexNet).
Increasing the height and width of the input image results in a higher-resolution input to the first FC layer.
At the CNN architecture level, an easy way around this is to design a CNN architecture that has global average pooling prior to the first FC layer -- this approach was popularized in the Network-in-Network (NiN) architecture~\cite{NiN}, and it is now used in other architectures such as SqueezeNet~\cite{SqueezeNet} and ResNet architectures~\cite{resnet}.
However, when using AlexNet or VGG19, the R-CNN authors developed a technique called {\em ROI Pooling} that allows any size input image to be used in concert with AlexNet/VGG FC layers.
ROI pooling is quite simple: no matter what the input image size is, ROI Pooling uses max-pooling to reshape the first FC layer's input to the size that it expects.

\subsection{Region Proposal Network in Faster R-CNN}

\begin{figure}[h]
\centering
\includegraphics[width=.9\textwidth]{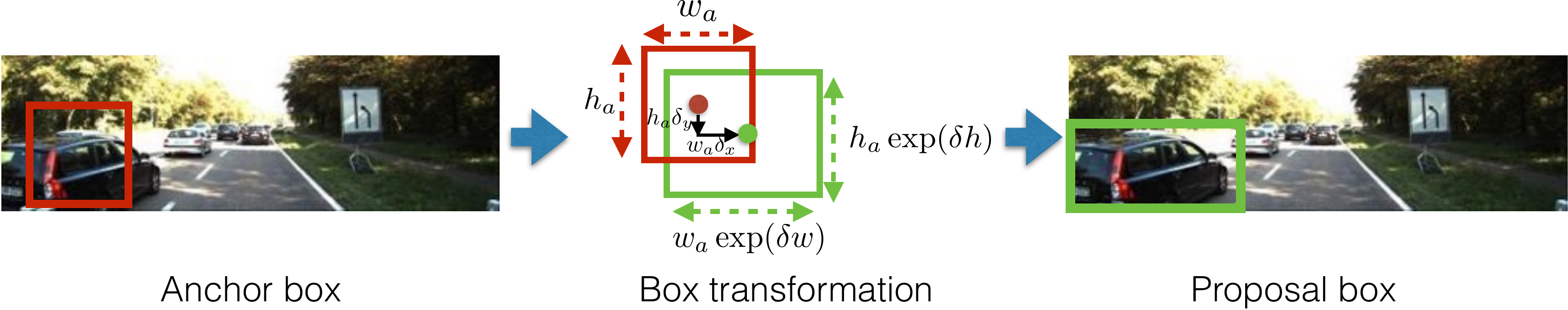}
\caption{Transformation from an anchor box (left) to a proposal (right). 4 relative coordinates are regressed by the RPN network to adjust the center position and the shape of the bounding box.}
\label{fig:BboxReg}
\end{figure}

We briefly review how the region proposal network (RPN) in Faster R-CNN generate proposals \cite{Faster-R-CNN} that will be useful later. RPN starts with convolution layers, which computes a high dimensional, low resolution feature map for the input image. Next, a small network slides through each spatial position in the feature map and generates rectangular region proposals centered around the position. Instead of computing the proposal's absolute coordinates, the RPN actually computes coordinates relative to a set of $k$ pre-selected reference boxes, or anchors. The transformation from an anchor to a proposal is illustrated in Fig. \ref{fig:BboxReg}. 

Intuitively, we want the anchors to be spatially close to the ground truth bounding boxes. In an extreme case, if an anchor box is too far away from the ground truth bounding box, learning to transform the anchor to the ground truth will be hopeless. Since anchors are centered at each spatial position on the feature map, and each position on the feature map corresponds to a patch of pixels on the original image, the resolution of the feature map affects the distance from a ground truth bounding box to its nearest anchor. In VGG16, for example, each position in conv5\_3 layer spatially corresponds to a $16 \times 16$ patch on the original image, so in the worst case, the nearest anchor to the center of a ground truth bounding box is $16\times \sqrt{2}/2 \approx 11.31$. As we will see later in this paper, reducing this distance, or relatively, increasing the ``anchor density'' will significantly increase the localization accuracy, thus improve the detection accuracy.

\section{Experimental Setup}
\label{sec:experimental-setup}

\subsection{Networks and training configuration} 
\label{sec:networks}

We train faster R-CNN networks built on the VGG16~\cite{VGG-19} and AlexNet~\cite{alexnet}, pretrained on the ImageNet-1k~\cite{imagenet} classification dataset. VGG16 has sixteen convolution layers ~\cite{VGG-19} and AlexNet has only five 
convolutional layers~\cite{alexnet}. Rather than using the convolution features of the last pooled layer (as is done in the original faster R-CNN paper), we use features from convolutional layers that are the bottom layers of the previous pooling layer. For VGG16 it is conv4\_3. We resize the $roi pooling$ window size accordingly. For VGG16 the window size changes from 7x7 to 13x13. We also reduce the feature stride by a factor of two for avoiding one pooling stage. For VGG16 and AlexNet, fully connected layers are used as the R-CNN branch. The weights in these layers are initialized with random gaussian noise. As the standard procedure introduced in faster R-CNN, we randomly sample 128 positive and 128 negative $roi$ proposals per batch to train the R-CNN layer. For all the experiments, we use initial learning rate of 0.0005, step size 50000 and momentum 0.9. A total of 70K iterations are run during R-CNN training starting from imagenet pre-trained weights for the convolution layers. 

\subsection{Dataset} 
\label{sec:dataset}

We use the KITTI object detection dataset. KITTI object has three categories that is car, pedestrian, bicyclists.
The dataset is annotated in three categories based on the occlusion and truncation of the objects. The hard category is heavily occluded and truncated whereas the easy category is relatively clearly visible. There are about 8000 images for both training and testing.
The moderate regime is used to rank
the competing methods in the benchmark. Our split of the KITTI train and validation sets (each containing half of the images) is the same as ~\cite{3DOP}. The evaluation criteria is the same that is prescribed in the KITTI development kit. In KITTI's evaluation criteria, proposal boxes having overlap with the ground truth or IoU greater than 70\% are counted as true detection for cars. 

The Faster R-CNN algorithm has been shown to deliver high accuracy on the PASCAL~\cite{PASCAL} dataset. Adapting that pipeline from PASCAL to KITTI poses a few natural challenges. First, the image sizes in the KITTI dataset is 1242x375 pixels whereas the image sizes in PASCAL dataset is 500 pixels in the longest dimension (many PASCAL images are 500x333 or 333x500). More importantly, the KITTI dataset contains heavily occluded and truncated objects. These objects come in multiple scales. The presence of objects at multiple scales make it difficult to attain high accuracy specially for small objects. 

\section{Input image resolution}
\label{sec:image_resolution}

We performed extensive design space search of Faster R-CNN configurations on the KITTI dataset. Our starting point is the VGG16 network that has achieved high accuracy in both image classification~\cite{VGG-19} and localization~\cite{Faster-R-CNN}. We performed an input image scaling experiment to find its impact on the accuracy. In these experiments, the shorter side of the KITTI images were fixed at 1295 pixels. In the Faster R-CNN codebase, the default off-the-shelf configuration resizes all images to 1000 pixels in the long dimension. 

KITTI images have a native resolution of 1295x375, so the default Faster R-CNN behavior is to resize KITTI images to 1000x302. But, is this resizing scheme ideal for obtaining high accuracy? To find out, we doubled the input image height and width to 2000x604. This has the effect of doubling the height and width of the activations (outputs) from all convolutional layers. For example, the conv5\_3 activations -- which serves as input to both the region proposal network (RPN) and the classification network -- double in height and width. With the image upsampled to 2000x604, we see in Table~\ref{T:VGG16_imageSize} that the KITTI car-detection accuracy increases for easy, medium, and hard by 7.1, 15.5, and 12.6 percentage points, respectively. In a world where half of a percentage point is considered significant, we can say with certainty that the input resolution has a {\em major} impact on accuracy.
  
Can further upsampling of the image lead to further improvements in accuracy?
We attempted to perform experiments with upsampling beyond 2000x604, but the volume of activation planes exceeded the 12GB of available memory on an {\em NVIDIA Titan X} GPU.
In the next section, we consider shallower networks with fewer layers of activation planes, which enables us to move to even higher input resolutions.


\begin{table}[t]
  \caption{KITTI car detection accuracy using different input image sizes to VGG16. In these experiments, we use conv5\_3 features from VGG16.}
  \label{T:VGG16_imageSize}
  \centering
   \begin{tabular}{llll}
   
    \toprule
    \multicolumn{4}{c}{AP}                   \\
    \cmidrule{2-4}
   
    Input resolution   & Easy     & Medium & High  \\
    \midrule
    1000x302 & 80.3  & 63.0    & 52.3  \\
    2000x604  & 87.4  & 78.5    & 64.9  \\
    \bottomrule
 
  \end{tabular}
\end{table}

\section{Shallow convolutional models}
\label{sec:shallow_models}

So far, we have upsampled the input image until we ran out of on-chip GPU memory when training R-CNN models with a VGG16-based feature representation.
Based on what we have seen so far, it seems that further upsampling may lead to further gains in accuracy.
We need to find a configuration that requires less memory for a given image size, and then we will exploit this extra memory to further upsample the image.
One idea would be to decrease the batch size to save memory, but we are already using a batch size of 1, and it's not clear how to reduce the batch size below 1.
Could we reduce the memory footprint by {\em reducing the number of layers} in the CNN?
Much of the recent literature shows that fewer layers in a CNN leads to lower accuracy (all else held equal).
But, our goal is to configure a CNN with fewer layers (and moderately lower accuracy) and then increase the input image resolution (leading to much higher accuracy).

To evaluate this idea, we configure R-CNN to use 2000x604 images (2x height and 2x width compared to our original starting point), using conv4\_3 instead of conv5\_3 features.
In VGG16's scheme for naming layers, conv4\_3 is the 10th layer, and conv5\_3 is the 13th layer in the CNN.
We expected that the accuracy of R-CNN with conv4\_3 would be slightly lower than R-CNN with conv5\_3, but as we show in Table~\ref{T:depth} that the accuracy is {\em higher} with conv4\_3 by 5.5, 9.4, and 12.4 percentage-points for easy, medium, and hard detections.
Cumulatively, the improvement in accuracy from conv5\_3 with 1000x302 images to conv4\_3 with 2000x604 images is a whopping 12.6, 24.9, and 25.0 percentage-points for easy, medium, and hard, respectively.

How does further reducing the CNN's depth affect accuracy?
We initially considered using the earlier layers of VGG16 as input to the Region Proposal Network.
But, earlier layers in VGG16 have been downsampled less, so their activations have a larger height and width.
We found that the off-the-shelf implementation of RPN comes to dominate the end-to-end computation time with very large height and width input grids.
Besides depth, one of the differences between VGG16 and AlexNet is that AlexNet downsamples more aggressively in the early layers -- for example AlexNet has stride=4 in the conv1 layer (4x downsampling), while the conv1 layer of VGG16 has stride=1 (no downsampling).
So, to evaluate this question of how using shallower ($<$10 conv layers) network impacts accuracy, we use AlexNet instead of VGG16.
We use conv5 (5th layer) activations as input to the R-CNN region-proposal and classification branches, and we report the results in Table~\ref{T:depth}.
With resolution of 2000x604 for both AlexNet-conv5 and VGG16-conv4\_3, the VGG16-based configuration delivers significantly higher accuracy on easy, medium, and hard detections in Table~\ref{T:depth}.
We have additional memory available when running AlexNet with 2000x604 input images, so we now try upsampling the AlexNet input images to 5000x1510. 
In this configuration, on the easy detections, AlexNet with 5000x1510 input is within 0.5 of a percentage-point of our best VGG16-based result so far.
On medium and hard categories, VGG16 conv4\_3 with an input resolution of 2000x604 delivers higher accuracy than AlexNet with 2000x604 or 5000x1510 input images.

\begin{figure}[h]
  \centering

 \includegraphics [width=3in]{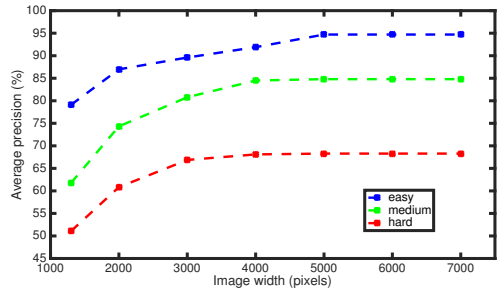} 
 \caption{Evaluation of how input resolution affects accuracy on KITTI car detection. In this configuration, we use AlexNet conv5 as input features to the R-CNN branches.}
\label{fig:ap_vs_scale}
\end{figure}

We also conduct a sweep of input image sizes applied to an AlexNet-based R-CNN model that uses conv5 features.
We show the results of this sweep in Figure~\ref{fig:ap_vs_scale}.
We observe that KITTI car detection accuracy steadily climbs from a baseline resolution of 1000x302 to a plateau at resolution 5000x1510.
Beyond this resolution, we have not observed further accuracy improvements at larger sizes such as 6000x1811 or 7000x2133.

\begin{table}[t]
  \caption{Impact of CNN depth on accuracy. Conventional wisdom would suggest that deeper representations would produce higher accuracy, but we find otherwise. AP numbers are for car detection on the KITTI dataset.}
  \label{T:depth}
  \centering
   \begin{tabular}{llllll}
   
    \toprule
    & & & \multicolumn{2}{c}{AP}                   \\
    \cmidrule{4-6}
   
    CNN Architecture & Conv layer name (depth) & Input image resolution & Easy     & Medium & Hard  \\
    \midrule
    VGG16 & conv5\_3 (13) & 2000x604 & 87.4 & 78.5 & 64.9 \\
    VGG16 & conv4\_3 (10) & 2000x604 & 92.9 & 87.9 & 77.3 \\
    AlexNet & conv5 (5) & 2000x604 & 86.7 & 71.6 & 56.1 \\
    AlexNet & conv5 (5) & 5000x1510 & 92.4 & 82.5 & 68.2  \\

    \bottomrule
 
  \end{tabular}
\end{table}

\section{Further improvements}
\label{sec:further-improvements}

\subsection{Context windows}

Chen et al.~\cite{3DOP} proposed {\em context windows} as a way to include information from adjacent pixels of a proposed bounding box. A context window is a bounding box that is scaled up from the original bounding box proposal of the RPN network. In the experiments with context window, in addition to the original R-CNN branch, an extra R-CNN branch is added that trains on the features extracted from context window. The original R-CNN features and the context R-CNN features are concatenated before classification. We add a context branch in addition to the usual R-CNN branch with a spatial bounding box scaling of 1.5. When applying the context window to an AlexNet-based R-CNN configuration, we find that the accuracy of all the categories improve as shown in Table \ref{T:overall-results}. The improvement is significant in small image sizes and provides diminishing returns as we scale up the input image size.

\subsection{Optimal Anchor-box shape}

\begin{figure}[h]

\hfill
\subfigure[Distribution of bounding box shapes of cars in the KITTI dataset]{\includegraphics[width=.4\linewidth]{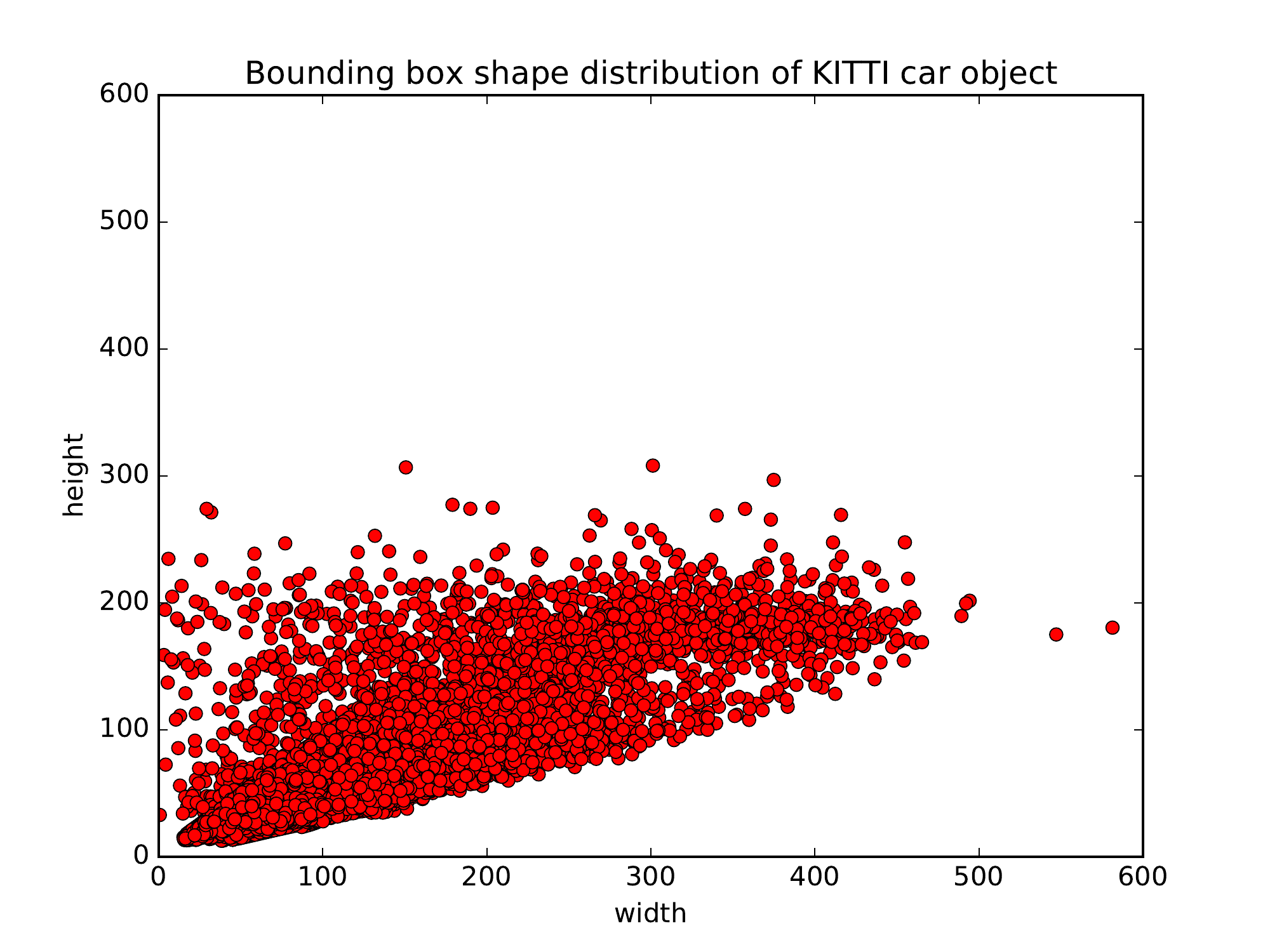}}
\hfill
\subfigure[9 Anchor box shapes (white crosses) selected by \textit{K-Means}]{\includegraphics[width=.4\linewidth]{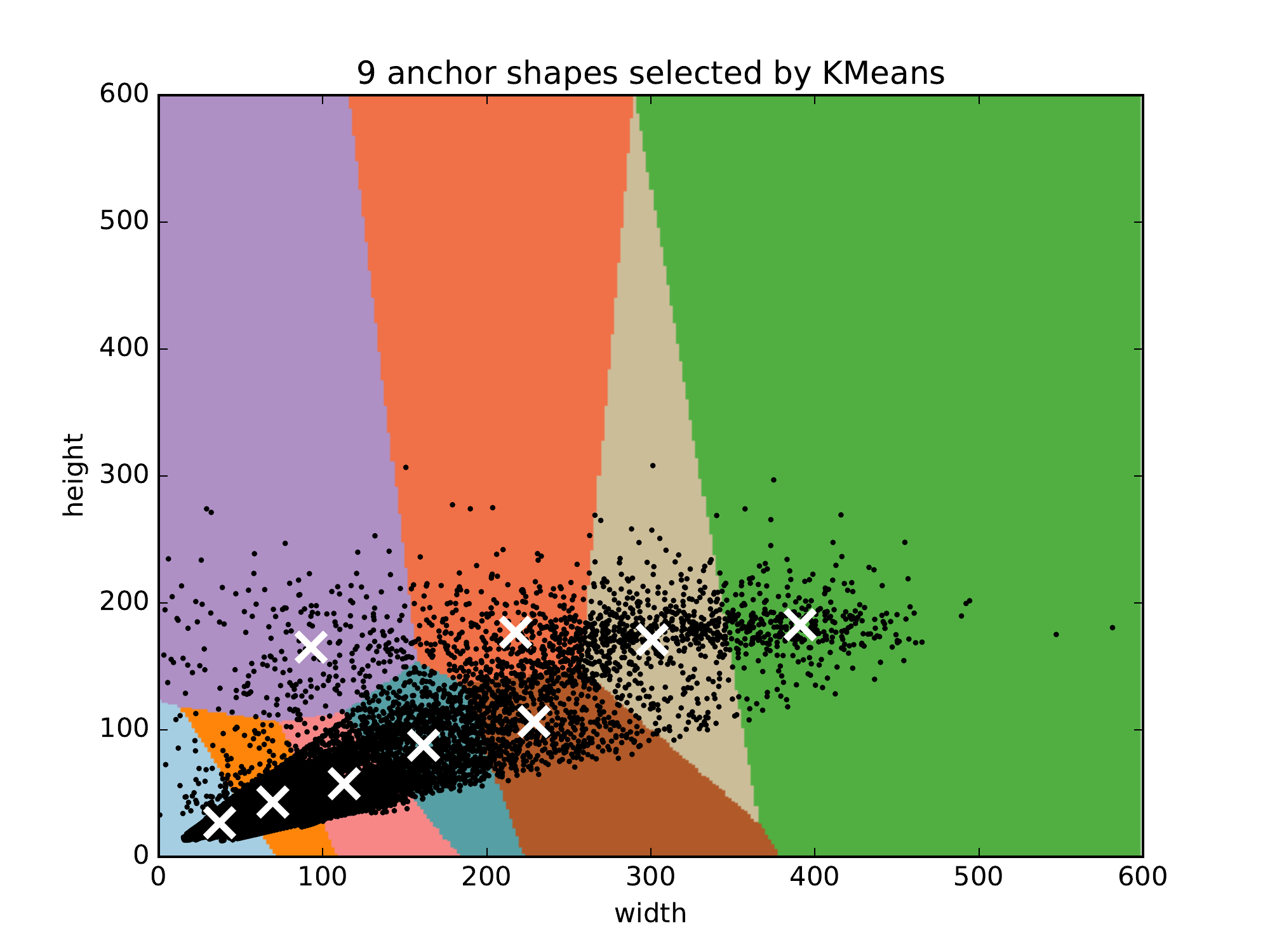}}
\hfill
\caption{Bounding box shape distribution of the car object in the KITTI dataset is plotted on the left. 9 anchor shapes computed by K-Means are plotted in the right figure as white crosses. }
\label{fig:bboxCar}
\end{figure}

In \cite{Faster-R-CNN}, default anchor shapes are arbitrarily chosen by reshaping a $16 \times 16$ square box by 3 scales and 3 aspect ratios. But is there a better way to choose anchor shapes for input images and target objects? Intuitively, we want the anchors to have similar shapes with the ground truth bounding boxes. The shape of a bounding box can be characterized by its width $w$ and its height $h$. The width and height distribution of the car object in the KITTI training data set is plotted in Fig. \ref{fig:bboxCar}(a). The problem of choosing the "most similar" $k$ anchor shapes can be formulated as the following: given a set of ground truth bounding box shape observations $\{(w_i, h_i)\}$, find $k$ anchors such that the sum of the distance (in the shape space of $(width, height)$) between each ground truth box to its nearest anchor is minimized. This problem can be effectively solved by \textit{K-means}. The optimal anchor shapes are plotted in Fig. \ref{fig:bboxCar}(b). These anchors are optimized specifically for the car category, but the idea of optimizing anchors by considering ground truth bounding box statistics can be generalized to multi-category object detection as well. 

We tested the AlexNet-based Faster R-CNN's detection accuracy with different anchor box selection schemes, and the result is shown in Table \ref{T:anchor-box}. We fix the number of anchors to be 9. From the bounding box shapes of cars in the training data set, we used K-Means to select 9 optimal anchor boxes. As comparison, we used a set of default anchors with 3 scales and 3 aspect ratios as in \cite{Faster-R-CNN}. When image width is $1242$, using the K-Means selected anchors improves AP significantly comparing with the default anchors. When the image width is $2500$, we could see that AP with default shapes are slightly better for easy and moderate category, but using K-Means selected anchors still improves AP for the hard category by 4 percentage points. As we further scale the image width to $5000$, the performance gain saturates, we still observe some improvement by using K-Means selected anchors.  We have not yet combined context windows with our anchor box improvements, but it is possible that this combination will yield a further improvement in accuracy.

\begin{table*}[htb]
	\footnotesize
    \caption{Impact of anchor box shape on accuracy. These results use AlexNet conv5 features.}
	\label{T:anchor-box}
	\centering
	\begin{tabulary}{17.2cm}{CCCCCC} 
	\hline
DNN architecture & Input resolution & Anchor shape selection scheme & \multicolumn{3}{c}{AP} \\
\cline{4-6}
&&&Easy & Moderate & Hard \\ \hline
AlexNet & 1242x375 & Default shape & 70.37 & 54.44 & 46.33 \\ 
AlexNet & 1242x375 & K-Means & 76.12 & 59.29 & 47.38 \\ 
AlexNet & 2500x755 & Default shape & 84.12 & 72.14 & 58.29 \\ 
AlexNet & 2500x755 & K-Means & 83.27 & 71.43 & 62.42 \\ 
AlexNet & 5000x1510 & Default shape & 91.33 & 84.52 & 69.90 \\ 
AlexNet & 5000x1510 & K-Means & \textbf{91.44} & \textbf{85.98} & \textbf{70.04} \\ \hline
	\end{tabulary}
\end{table*}

\begin{table}[h]
  \caption{Summary of results on KITTI~\cite{kittiData} car detection. All of our results are based on Faster R-CNN. To our knowledge, all of the related work discussed in this table also uses a version of R-CNN.}
  \label{T:overall-results}
  \footnotesize
  \centering
  \begin{tabulary}{\textwidth}{CCCCCCCCCCCCCCCC} 
      & & & & & & AP & \\
   
    
     Source & CNN Architecture & Feature layer (Depth) & Input resolution & Context window & Easy & Medium & Hard\\

     SDP+CRC~\cite{SDP_CRC} & VGG16 & {\tiny conv{3\_3, 4\_3, 5\_3} (7,10,13)} & multiple  & no & 90.3 & 83.5 & 71.1 \\ 
     Mono3D~\cite{mono3D} & VGG16 & conv5\_3 (13) & {\tiny not reported} & yes & 92.3 & {\bf 88.7} & {\bf 79.0} \\ 

\vspace{0.1cm} \\
     
     ours & VGG16 & conv5\_3 (13) & 1000X302 &  no & 80.25  & 62.96    & 52.3 \\ 
     ours & VGG16 & conv5\_3 (13)  & 2000X604 & no & 87.35  & 78.49    & 64.93 \\ 
     ours & VGG16 & conv4\_3 (10) & 2000X604  & no & 92.9 & 87.9 & 77.3 \\ 
     ours & VGG16 & conv4\_3 (10) & 5000x1510 & no & \multicolumn{3}{c}{out of memory} \\ 
\vspace{0.1cm} \\ 
     ours & AlexNet & conv5 (5) & 1000X302 & no & 67.5 &	49.44 &	38.9 \\
     ours & AlexNet & conv5 (5) & 2000X604 & no & 86.7 &	71.6	& 58.1 \\
     ours & AlexNet & conv5 (5) & 5000x1510 & no & 92.4 & 82.5 & 68.2\\ 
\vspace{0.1cm} \\ 
     ours & AlexNet & conv5 (5) & 1000X302 & yes & 71.58 &	51.13 &	40.9 \\ 
     ours & AlexNet & conv5 (5) & 2000X604 & yes & 86.98 & 74.32 & 60.83  \\ 
     ours & AlexNet & conv5 (5) & 5000x1510 & yes & {\bf 94.7} & 84.8 & 68.3\\ 


  \end{tabulary}
\end{table}

\section{Discussion} 
\label{sec:discussion}

\begin{figure}[h]

\hfill
\subfigure[Precision-recall curve for the AlexNet network with context window. The input image size is 5000. The conv5 features are used in this experiment.]{\includegraphics[width=.45\linewidth]{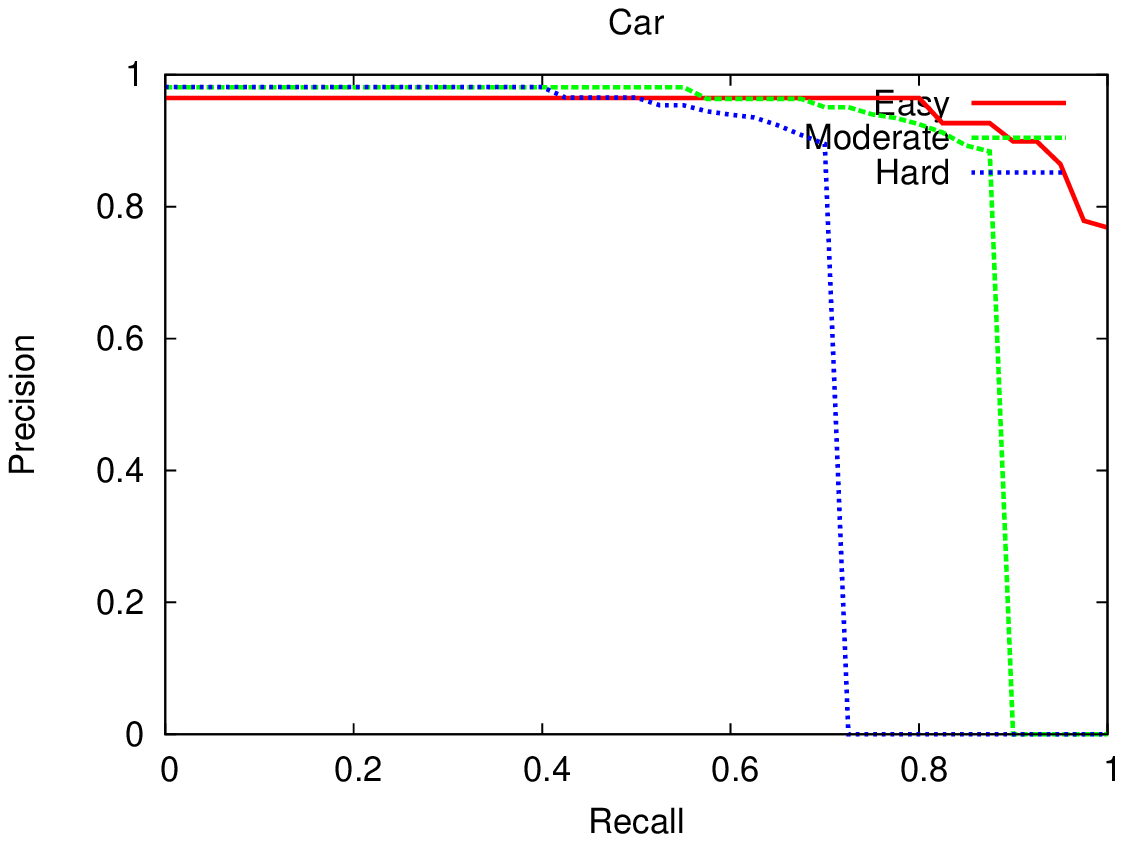}}
\hfill
\subfigure[Precision-recall curve for the VGG16 network without context window. The input image size is 2000. The conv4\_3 features are used in this experiment.]{\includegraphics[width=.4\linewidth]{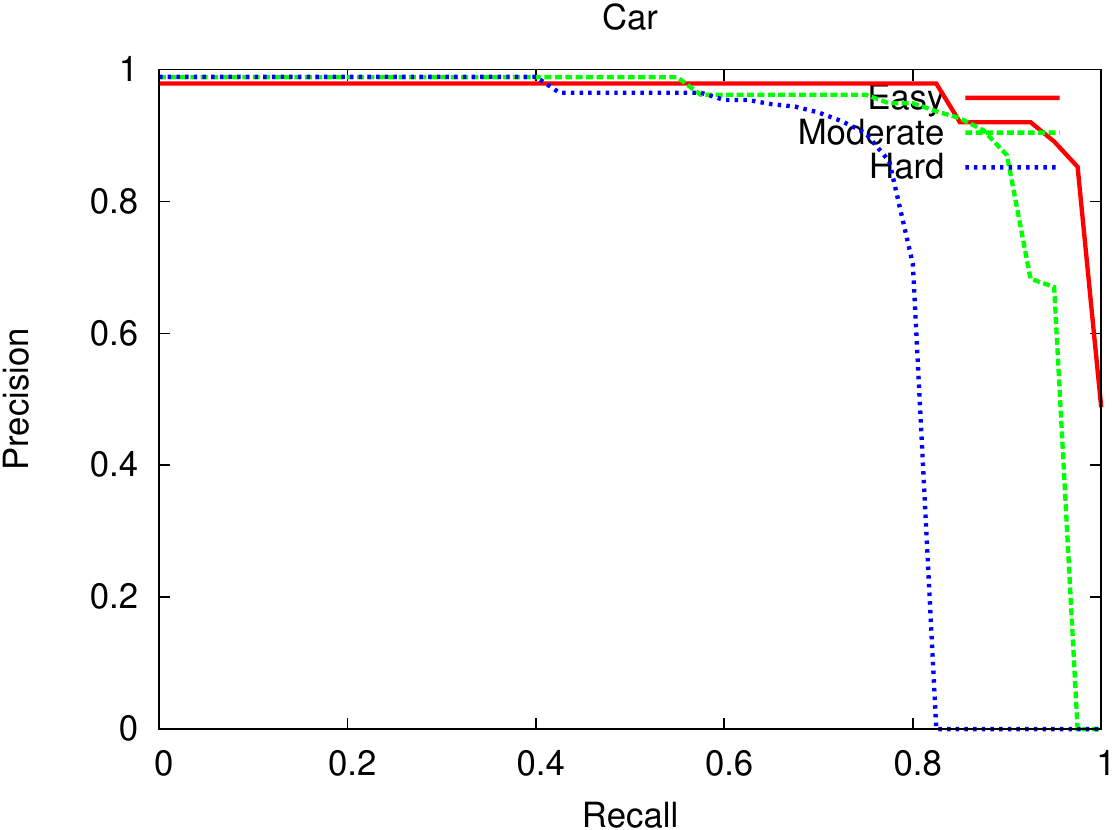}}
\hfill
\caption{Precision vs. recall curve for KITTI's car detection system.}
\label{fig:apVsRecall}
\end{figure}

\begin{figure}[h!]

\hfill
\subfigure[Success.]{\includegraphics[width=.3\linewidth]{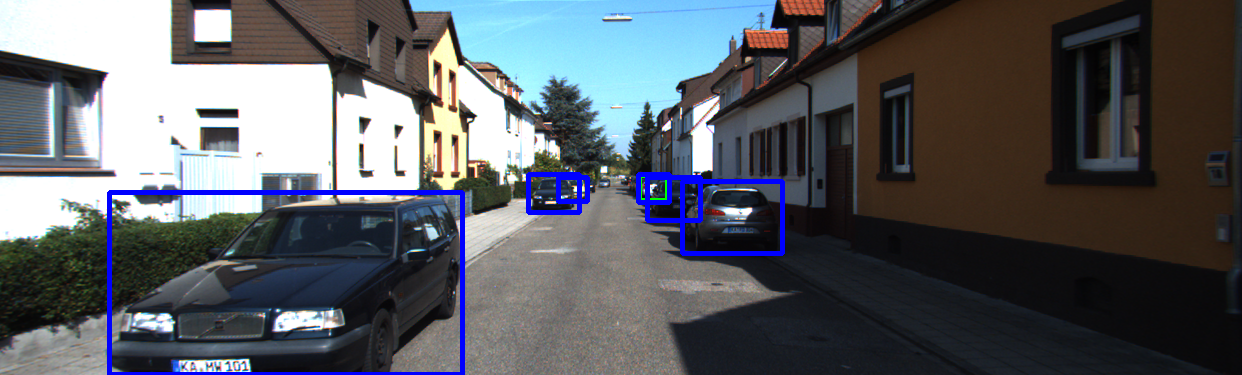}}
\hfill
\subfigure[Predicting one box for two adjacent cars.]{\includegraphics[width=.3\linewidth]{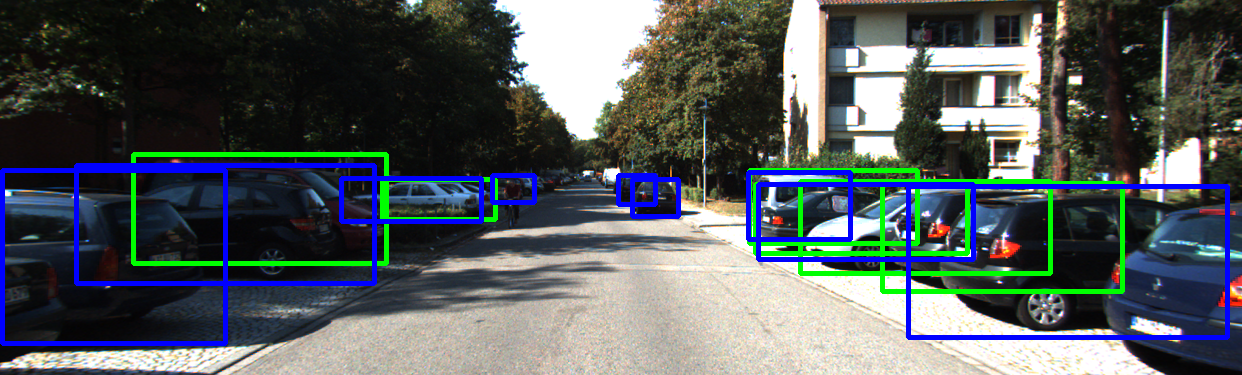}}
\hfill
\subfigure[Failure]{\includegraphics[width=.3\linewidth]{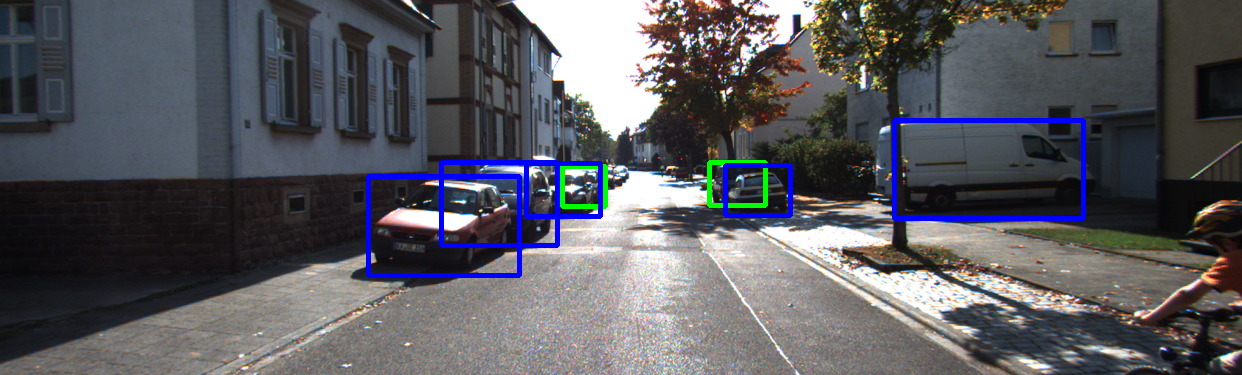}}
\hfill
\caption{Examples of success(a) and failure modes(b,c) in the hard category using AlexNet $conv5$ with a context window and input resolution of 5000x1510 pixels. The green boxes show the ground truth label of heavily truncated car. In (a), the model successfully predict a highly occluded car. In (b), the predicted bounding box encompass two cars that are adjacent to each other. In (c), the predicted bounding box enclose a visible car but completely misses the car that is truncated.}
 \label{fig:example_hard}
\end{figure}

We show the precision-recall curve for our KITTI car detection in Fig. \ref{fig:apVsRecall}(a). We used AlexNet-conv5 with a context window and input image resolution of 5000x1510, as in the final row of Table~\ref{T:overall-results}. We observe that the precision of the easy category is very high even at very high recall. However, the precision in the hard category suffers at high recall. Improving the precision of results on the hard category with our method will be a target of future work. We show a few examples of success and failure modes in the hard category in Fig \ref{fig:example_hard}. In Fig \ref{fig:example_hard}(a), the model successfully predict a highly occluded car while in Fig \ref{fig:example_hard}(b) the predicted bounding box encompass two cars that are adjacent to each other. In Fig \ref{fig:example_hard}(c), the predicted bounding box enclose a visible car but completely misses the car that is truncated. The precision-recall curve using the conv4\_3 features of VGG16 is shown in Fig.\ref{fig:apVsRecall}(b). The precision at high recall for the hard category improves significantly. The inference time using AlexNet conv5 layer with input image size of 5000x1510 and VGG16 conv4\_3 layer with input image size of 2000x604 is 0.34s and 0.6s respectively. Inference times for other published high accuracy methods on the KITTI dataset are 3s for 3DOP~\cite{3DOP} and 0.4s for SDP+CRC~\cite{SDP_CRC}.

\section{Conclusions}
\label{sec:conclusions}

In summary, we have shown that shallow networks perform well in achieving high accuracy on detecting cars in the road. We have shown that input image resolution has a large impact on the accuracy of car detection using the faster R-CNN network. For very deep models, shallow layers can (surprisingly) provide higher accuracy than the later convolutional layers deep in the network. Shallow models like AlexNet can achieve high accuracy when the input image is upsampled. In addition, we have used an anchor box selection method and context window to further enhance  car detection accuracy. 
We believe that our findings will inspire the research community to evaluate shallow models for achieving high accuracy on object detection tasks.

\section*{Acknowledgments}
Khalid Ashraf was supported by the National Science Foundation under Award number 125127. We thank Kostadin Ilov for many help with computational hardware. Thanks to Ross Girshick for comments on some initial results. Thanks to Fan Yang and Kaustav Kundu for clarification on their results.

{\small
\bibliographystyle{ieee}
\bibliography{bibliography}

\begin{thebibliography}{10}\itemsep=-1pt

\bibitem{deepDriving}
C.~Chen, A.~Seff, A.~Kornhauser, and J.~Xiao.
\newblock Deepdriving: Learning affordance for direct perception in autonomous
  driving.
\newblock In {\em CVPR}, 2015.

\bibitem{mono3D}
X.~Chen, K.~Kundu, Z.~Zhang, H.~Ma, S.~Fidler, and R.~Urtasun.
\newblock Monocular 3d object detection for autonomous driving.
\newblock In {\em CVPR}, 2016.

\bibitem{3DOP}
X.~Chen, K.~Kundu, Y.~Zhu, A.~Berneshawi, H.~Ma, S.~Fidler, and R.~Urtasun.
\newblock 3d object proposals for accurate object class detection.
\newblock {\em NIPS}, 2015.

\bibitem{imagenet}
J.~Deng, W.~Dong, R.~Socher, L.-J. Li, K.~Li, and L.~Fei-Fei.
\newblock {ImageNet}: A large-scale hierarchical image database.
\newblock In {\em CVPR}, 2009.

\bibitem{multibox-1}
D.~Erhan, C.~Szegedy, A.~Toshev, and D.~Anguelov.
\newblock Scalable object detection using deep neural networks.
\newblock In {\em CVPR}, 2014.

\bibitem{PASCAL}
M.~Everingham, L.~V. Gool, C.~K.~I. Williams, J.~Winn, and A.~Zisserman.
\newblock The pascal visual object classes (voc) challenge.
\newblock {\em International Journal of Computer Vision (IJCV)}, 2010.

\bibitem{FelzenszwalbPAMI10}
P.~Felzenszwalb, R.~Girshick, D.~McAllester, and D.~Ramanan.
\newblock {Object Detection with Discriminatively Trained Part Based Models}.
\newblock {\em PAMI}, 2010.

\bibitem{kittiData}
A.~Geiger, P.~Lenz, and R.~Urtasun.
\newblock Are we ready for autonomous driving? the kitti vision benchmark
  suite.
\newblock In {\em CVPR}, 2012.

\bibitem{Fast-R-CNN}
R.~Girshick.
\newblock Fast r-cnn.
\newblock In {\em ICCV}, 2015.

\bibitem{R-CNN}
R.~B. Girshick, J.~Donahue, T.~Darrell, and J.~Malik.
\newblock Rich feature hierarchies for accurate object detection and semantic
  segmentation.
\newblock In {\em CVPR}, 2014.

\bibitem{SPPnet}
K.~He, X.~Zhang, S.~Ren, and J.~Sun.
\newblock Spatial pyramid pooling in deep convolutional networks for visual
  recognition.
\newblock {\em arXiv:1406.4729}, 2014.

\bibitem{resnet}
K.~He, X.~Zhang, S.~Ren, and J.~Sun.
\newblock Deep residual learning for image recognition.
\newblock {\em arXiv:1512.03385}, 2015.

\bibitem{hillel2012}
A.~B. Hillel, R.~Lerner, D.~Levi, and G.~Raz.
\newblock Recent progress in road and lane detection: a survey.
\newblock {\em Machine Vision and Applications}, 2012.

\bibitem{huval2015}
B.~Huval, T.~Wang, S.~Tandon, J.~Kiske, W.~Song, J.~Pazhayampallil,
  M.~Andriluka, P.~Rajpurkar, T.~Migimatsu, R.~Cheng-Yue, F.~Mujica, A.~Coates,
  and A.~Y. Ng.
\newblock An empirical evaluation of deep learning on highway driving.
\newblock {\em arXiv preprint arXiv:1504.01716v3}, 2015.

\bibitem{SqueezeNet}
F.~N. Iandola, M.~W. Moskewicz, K.~Ashraf, S.~Han, W.~J. Dally, and K.~Keutzer.
\newblock Squeezenet: Alexnet-level accuracy with 50x fewer parameters and
  $<$1mb model size.
\newblock {\em arXiv:1602.07360}, 2016.

\bibitem{alexnet}
A.~Krizhevsky, I.~Sutskever, and G.~E. Hinton.
\newblock {ImageNet Classification with Deep Convolutional Neural Networks}.
\newblock In {\em NIPS}, 2012.

\bibitem{NiN}
M.~Lin, Q.~Chen, and S.~Yan.
\newblock Network in network.
\newblock {\em arXiv:1312.4400}, 2013.

\bibitem{rajpurkar2015}
P.~Rajpurkar, T.~Migimatsu, J.~Kiske, R.~Cheng-Yue, S.~Tandon, T.~Wang, and
  A.~Ng.
\newblock Driverseat: Crowdstrapping learning tasks for autonomous driving.
\newblock {\em arXiv preprint arXiv:1512.01872v1}, 2015.

\bibitem{Faster-R-CNN}
S.~Ren, K.~He, R.~Girshick, and J.~Sun.
\newblock Faster r-cnn: Towards real-time object detection with region proposal
  networks.
\newblock In {\em NIPS}, 2015.

\bibitem{overfeat}
P.~Sermanet, D.~Eigen, X.~Zhang, M.~Mathieu, R.~Fergus, and Y.~LeCun.
\newblock Overfeat: Integrated recognition, localization and detection using
  convolutional networks.
\newblock In {\em ICLR}, 2014.

\bibitem{VGG-19}
K.~Simonyan and A.~Zisserman.
\newblock Very deep convolutional networks for large-scale image recognition.
\newblock {\em arXiv:1409.1556}, 2014.

\bibitem{googlenet}
C.~Szegedy, W.~Liu, Y.~Jia, P.~Sermanet, S.~Reed, D.~Anguelov, D.~Erhan,
  V.~Vanhoucke, and A.~Rabinovich.
\newblock Going deeper with convolutions.
\newblock {\em arXiv:1409.4842}, 2014.

\bibitem{multibox-2}
C.~Szegedy, S.~Reed, D.~Erhan, , and D.~Anguelov.
\newblock Scalable, high-quality object detection.
\newblock {\em arXiv:1412.1441 (v1)}, 2015.

\bibitem{xiang2015}
Y.~Xiang, W.~Choi, Y.~Lin, and S.~Savarese.
\newblock Data-driven 3d voxel patterns for object category recognition.
\newblock In {\em CVPR}, 2015.

\bibitem{SDP_CRC}
F.~Yang, W.~Choi, and Y.~Lin.
\newblock Exploit all the layers: Fast and accurate cnn object detector with
  scale dependent pooling and cascaded rejection classifiers.
\newblock In {\em CVPR}, 2016.

\bibitem{segDeepM}
Y.~Zhu, R.~Urtasun, R.~Salakhutdinov, and S.~Fidler.
\newblock segdeepm: Exploiting segmentation and context in deep neural networks
  for object detection.
\newblock In {\em CVPR}, 2015.

\end{thebibliography}
}

\end{document}